\documentclass[letterpaper, 10 pt, conference]{ieeeconf}  % Comment this line out

\IEEEoverridecommandlockouts                              % This command is only
\usepackage[colorlinks,linkcolor=green,citecolor=red,urlcolor=blue,bookmarks=true,hypertexnames=true]{hyperref} 
\usepackage[utf8]{inputenc}
\usepackage[T1]{fontenc}
\usepackage{graphicx} % for pdf, bitmapped graphics files
\usepackage{epsfig} % for postscript graphics files
\usepackage{mathptmx} % assumes new font selection scheme installed
\usepackage{mathptmx} % assumes new font selection scheme installed
\usepackage{amsmath} % assumes amsmath package installed
\usepackage{amssymb}  % assumes amsmath package installed
\usepackage{booktabs}
\usepackage{url}
\usepackage{multirow}
\usepackage{multicol}
\usepackage{threeparttable} % 引入threeparttable宏包
\usepackage{balance}

\title{\LARGE \bf Learning to Generate Vectorized Maps at Intersections with Multiple Roadside Cameras}

\author{Quanxin Zheng$^{2}$, Miao Fan$^{1,*}$, Shengtong Xu$^{3}$, Linghe Kong$^{4}$, Haoyi Xiong$^{5}$ % <-this % stops a space
\\
\\
$^{1}$Beijing Institute of Graphic Communication,
$^{2}$NavInfo Co. Ltd.,
$^{3}$Autohome Inc.,
\\
$^{4}$Shanghai Jiaotong University,
$^{5}$Baidu Inc.
\thanks{*Corresponding author: Miao Fan (miao.fan@ieee.org), professor at Beijing Institute of Graphic Communication, senior member of IEEE.}
}

\begin{document}

\maketitle
\thispagestyle{empty}
\pagestyle{empty}

%%%%%%%%%%%%%%%%%%%%%%%%%%%%%%%%%%%%%%%%%%%%%%%%%%%%%%%%%%%%%%%%%%%%%%%%%%%%%%%%
\begin{abstract}
Vectorized maps are indispensable for precise navigation and the safe operation of autonomous vehicles. Traditional methods for constructing these maps fall into two categories: offline techniques, which rely on expensive, labor-intensive LiDAR data collection and manual annotation, and online approaches that use onboard cameras to reduce costs but suffer from limited performance, especially at complex intersections. To bridge this gap, we introduce the \underline{M}ultiple \underline{R}oadside \underline{C}amera-based \underline{V}ectorized \underline{Map} approach, MRC-VMap, a cost-effective, vision-centric, end-to-end neural network designed to generate high-definition vectorized maps directly at intersections. Leveraging existing roadside surveillance cameras, MRC-VMap directly converts time-aligned, multi-directional images into vectorized map representations. This integrated solution lowers the need for additional intermediate modules--such as separate feature extraction and Bird’s-Eye View (BEV) conversion steps--thus reducing both computational overhead and error propagation. Moreover, the use of multiple camera views enhances mapping completeness, mitigates occlusions, and provides robust performance under practical deployment constraints. Extensive experiments conducted on $4,000$ intersections across $4$ major metropolitan areas in China demonstrate that MRC-VMap not only outperforms state-of-the-art online methods but also achieves accuracy comparable to high-cost LiDAR-based approaches, thereby offering a scalable and efficient solution for modern autonomous navigation systems.

\end{abstract}
{\keywords vectorized maps, roadside cameras, generative neural networks, edge computing}
%%%%%%%%%%%%%%%%%%%%%%%%%%%%%%%%%%%%%%%%%%%%%%%%%%%%%%%%%%%%%%%%%%%%%%%%%%%%%%%%

\section{Introduction}
%Vectorized maps serve as a cornerstone for autonomous driving systems, providing centimeter-level details regarding traffic elements, vectorized topology, and navigational information. These maps are indispensable for enabling precise navigation and ensuring the safe operation of autonomous vehicles, as they offer essential insights into road layouts, traffic signs, and other vital components. Such information facilitates the accurate localization of vehicles and enables them to anticipate upcoming conditions, which is crucial for informed real-time decision-making.

%Traditional methodologies for constructing vector maps predominantly rely on labor-intensive data collection through specialized survey vehicles equipped with LiDAR and cameras. This conventional approach involves intricate alignment processes, manual annotation, and offline data processing, which contribute to substantial costs and prolonged update cycles due to its dependence on labor-intensive manual efforts and limited scalability.

Vectorized maps serve as a cornerstone for autonomous driving systems, providing centimeter-level details regarding traffic elements, vectorized topology, and navigational information~\cite{zhang2014loam}. These maps are indispensable for enabling precise navigation and ensuring the safe operation of autonomous vehicles, as they offer essential insights into road layouts, traffic signs, and other vital components. Traditional methodologies for constructing vector maps predominantly rely on labor-intensive data collection through specialized survey vehicles equipped with LiDAR and cameras~\cite{shan2018lego}. This conventional approach involves sophisticated alignment processes, manual annotation, and offline data processing, which contribute to substantial costs and prolonged update cycles due to its dependence on labor-intensive manual efforts and limited scalability~\cite{shan2020lio}.

Recent research has explored online mapping methods using onboard sensors—primarily cameras and LiDAR—to generate vector maps in real time. Existing solutions, such as HDMapNet~\cite{li2022hdmapnet}, VectorMapNet~\cite{liu2023vectormapnet}, and MapTR~\cite{maptr} demonstrate the potential of multimodal data integration. However, these techniques suffer from restricted fields of view and occlusions, particularly at complex intersections, often leading to incomplete maps. Alternatively, VI-Map~\cite{vi-map} builds a small-scale real-world dataset by combining annotated camera images with LiDAR data. Despite achieving state-of-the-art results at challenging intersections, this method demands significant resources for aligning LiDAR and camera data, limiting its practicality for widespread use.

%Recent research has explored online mapping methods that exploit onboard sensors—specifically cameras and LiDAR—to generate maps in real time. Methods such as HDMapNet~\cite{li2022hdmapnet}, VectorMapNet~\cite{liu2023vectormapnet}, and MapTR~\cite{maptr} demonstrate the promise of integrating multimodal sensor data for vector map generation. However, these methods are limited by restricted fields of view and occlusions in complex intersection scenarios, leading to incomplete mapping data. In the meanwhile, VI-Map~\cite{vi-map} proposed to construct a small-scale, real-world dataset combining annotated camera images and LiDAR data. Although VI-Map achieved state-of-the-art results at complex intersections, it relies on intensive resource allocation for aligning LiDAR and camera data across various scenes, making widespread deployment impractical.

To the end, at least two technical challenges should be tackled to achieve our goals.
\begin{itemize}
    \item \textbf{Single versus Multiple:} While single camera systems fail to capture complete intersection details, converting diverse perspectives from multiple roadside cameras to a unified bird’s-eye view could meet our goal but also requires precise calibration, with errors propagating through the mapping process.

    \item \textbf{Precision versus Complexity:} High-precision vectorized mapping traditionally relies on computationally intensive multi-stage pipelines, where errors cascade through intermediate modules, compromising overall accuracy. Moreover, the computational demands of real-time high-definition map generation pose significant challenges for deployment on cost-sensitive edge devices, necessitating a more efficient approach.
    %High-precision vectorized mapping relies on the intensive computation with multi-stage pipelines exacerbate inaccuracies by compounding errors from intermediate modules. Real-time high-definition map generation demands efficient computation, challenging deployment on cost-sensitive edge devices.
\end{itemize}

To address these challenges, the present study introduces MRC-VMap, an end-to-end neural framework that efficiently generates precise vector maps of intersections through the use of roadside cameras positioned in multiple directions. By leveraging the pre-existing traffic surveillance infrastructure, MRC-VMap makes itself a practical, scalable, and cost-effective solution. In addition, MRC-VMap could be implemented using commodity AI chips for edge computing, such as the Jetson AGX Orin, achieving high-definition map generation with reduced computational overhead.

The main contributions of this study are as follows.

\begin{figure}
    \centering
  \includegraphics[width=0.9\columnwidth]{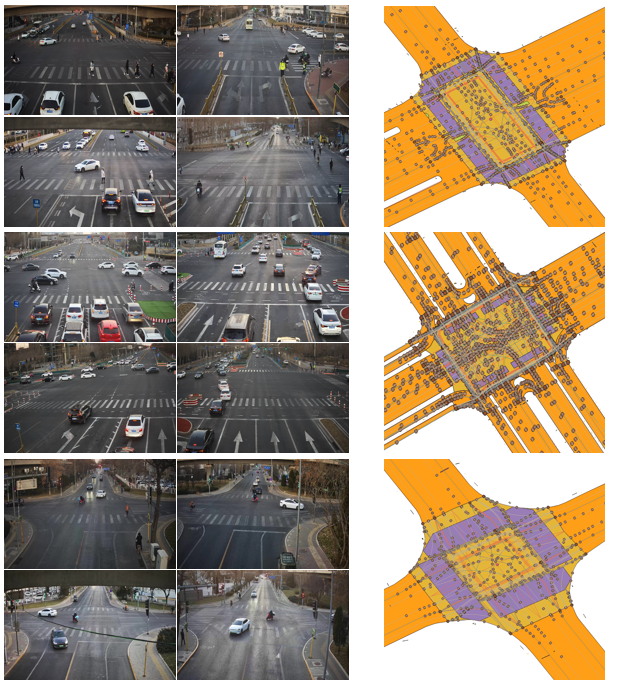}
  \caption{This figure presents a data sample from the Navinfo dataset, comprising three groups, with each group containing four images along with the corresponding vector ground truth data for the intersection. On the left side, the collected images, characterized by a resolution of 2 million pixels, are displayed, while the right side features the vector rendering of the intersection.}
  \label{fig:navinfo}
  \vspace{-5mm}
\end{figure}

\begin{itemize}

\item MRC-VMap utilizes time-synchronized images captured by multiple roadside cameras as input, directly producing a vector map of the intersection without necessitating any external parameters. The data fusion from multiple cameras significantly enhances the completeness of the vector map while lowering occlusions. 

\item To facilitate this study, we have compiled the Navinfo dataset in Fig.~\ref{fig:navinfo}, which comprises camera data and ground truth for 15 key mapping elements across 4,000 intersections in four major cities: Beijing, Shanghai, Guangzhou, and Shenzhen. This dataset is approximately 200 times larger than existing comparable dataset and includes critical elements such as pedestrian crossings, lane dividers, and road boundaries.

\item Empirical evaluations show that MRC-VMap not only surpasses the performance of existing online mapping methods but also achieves results that are comparable to traditional offline methods reliant on LiDAR technology, marking a significant advancement in the field of vector map generation for autonomous driving systems.

\end{itemize}

\section{Related Work}
Traditional vectorized map construction relies on labor-intensive data collection using surveying vehicles, where LiDAR point clouds and images are manually annotated. Techniques such as HDMapNet~\cite{li2022hdmapnet} exemplify this approach, which results in high operational costs and slow update cycles due to the need for heuristic post-processing and manual intervention. Recent research, however, has shifted towards end-to-end mapping generation technologies that minimize manual annotation and improve scalability. For example, VectorMapNet~\cite{liu2023vectormapnet} integrates surround-view camera images with LiDAR point clouds to predict sparse polygonal lines in a bird’s-eye view (BEV), while MapTR~\cite{maptr} employs Transformer networks that leverage only image inputs for effective positioning and classification through instance-level and point-level matching. Despite these advancements, onboard mapping continues to face challenges from limited fields of view and occlusions, which compromise map integrity, particularly at intersections. To mitigate these issues, VI-Map~\cite{vi-map} utilizes roadside infrastructure and LiDAR-based systems for enhanced mapping; however, the high installation costs limit its practicality.

\begin{figure*}[!htp]
    \centering
  \includegraphics[width=0.9\textwidth]{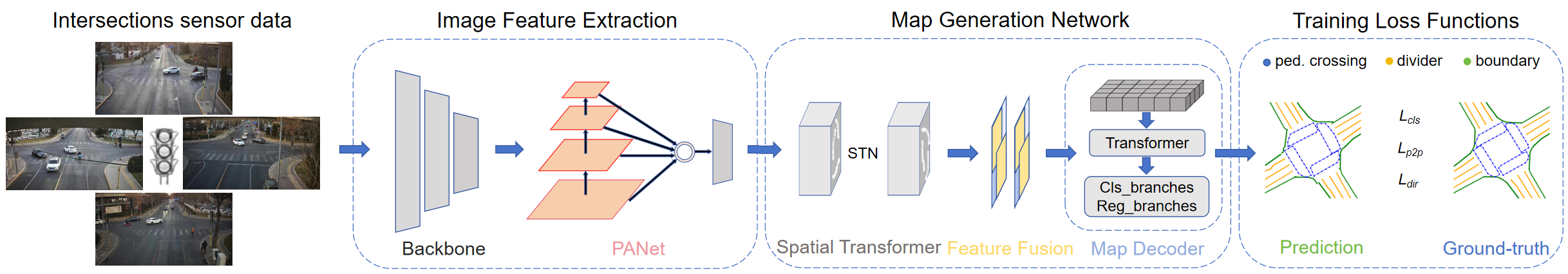}
  
  \caption{The pipeline of The MRC-VMap consists of three main parts: Image Feature Extraction, Map Generation Network, and Training Loss Functions. Among them, Image Feature Extraction includes processing by the Backbone network and the Neck (FPN) network. The Map Generation Network includes the Spatial Transformer Network (STN), the Feature Fusion Network,and the Output Head. The Training Loss Functions include $L_{\text{cls}}$ , $L_{\text{p2p}}$ , and $L_{\text{dir}}$.}
  \label{fig:lct}\vspace{-5mm}
\end{figure*}

\section{Problem Formulation}
We consider the \emph{problem of mapping multi-view images and their corresponding spatial transformations to a unified vectorized map}. Let the end-to-end mapping be denoted by the function $\mathcal{F}$ with parameters $\Theta$. Formally, given the input images 
\begin{equation}\label{eq2}
   \mathcal{X} = (\mathbf{X}_{1},\mathbf{X}_{2},\mathbf{X}_{3},\mathbf{X}_{4}), 
\end{equation}
where each $\mathbf{X}_i \in \mathbb{R}^{h \times w \times c}$ represents an image captured from a roadside camera, and given the spatial transformation matrices
\begin{equation}
    \mathcal{C} = (\mathbf{C}_{1},\mathbf{C}_{2},\mathbf{C}_{3},\mathbf{C}_{4}),
\end{equation}
with each $\mathbf{C}_i \in \mathbb{R}^{4 \times 4}$ mapping the corresponding camera view to the bird’s-eye view (BEV) feature space, the problem is to model the mapping
\begin{equation}\label{eq1}
    \mathcal{Y} = \mathcal{F}(\mathcal{X},\mathcal{C}; \Theta),
\end{equation}
where $\mathcal{Y}$ represents the output vectorized map. This formulation encapsulates the challenge of determining an appropriate function $\mathcal{F}$, parameterized by $\Theta$, that accurately transforms the inputs $\mathcal{X}$ and $\mathcal{C}$ into a high-fidelity vector map.

Due to the potential calibration errors and computational challenges of deriving BEV features from camera parameters, our work assumes there exits a BEV-Learner that autonomously learns the mapping from perspective view to BEV feature space. This setting by-passes the need for explicit camera parameters, succinctly reducing the original problem in Equation~\ref{eq1} into:
\begin{equation}
\label{eq1}
\mathcal{Y} = \mathcal{F}(\mathcal{X}; \Theta).
\end{equation}
Thus, our work intends to design the function $\mathcal{F}$ as a neural network for map generation and learn $\Theta$ from massive data.

%MRC-VMap is conceived as an end-to-end neural network architecture, denoted by the function $\mathcal{F}$, The parameters of this network are represented by $\Theta$. Thus, the notation $\mathcal{F}$($\Theta$) encapsulates the complete architecture of MRC-VMap.  Furthermore, we define $\mathcal{X}$ as the input image from the roadside cameras,  $\mathcal{C}$ as the transformation matrix parameters that map the current camera viewpoint to the BEV feature space, and $\mathcal{Y}$ as the output vector map, then MRC-VMap can be formulated as follows:

%\begin{equation}
%\label{eq1}
%\mathcal{Y} = \mathcal{F}(\mathcal{X},\mathcal{C}; \Theta).
%\end{equation}
%The variables $\mathcal{X}$ and $\mathcal{C}$ can be expressed as:
%\begin{equation}
%\label{eq2}
%\mathcal{X} = (\textbf{X}_{1},\textbf{X}_{2},\textbf{X}_{3},\textbf{X}_{4})
%\end{equation}
%and
%\begin{equation}
%\mathcal{C} = (\textbf{C}_{1},\textbf{C}_{2},\textbf{C}_{3},\textbf{C}_{4})
%\end{equation}

%Here, $\textbf{X}_1$,$\textbf{X}_2$,$\textbf{X}_3$,$\textbf{X}_4 \in \mathbb{R}^{h \times w \times c}$ represent the four images, each with a width of $w$ pixels, a height of $h$ pixels, and $c$ channels. The matrices $\textbf{C}_1$,$\textbf{C}_2$,$\textbf{C}_3$,$\textbf{C}_4$  denote the four $4*4$ spatial transformation matrices.

\section{Methodology}
To solve the formulated problem above, we propose MRC-VMap. The overall framework of MRC-VMap is illustrated in Fig.~\ref{fig:lct}, consisting of three major steps: \emph{image feature extraction}, \emph{map generation network}, and \emph{training loss function}. We introduce each key component as follows.

\subsection{Image Feature Extraction}
\begin{figure}
    \centering
  \includegraphics[width=0.5\textwidth]{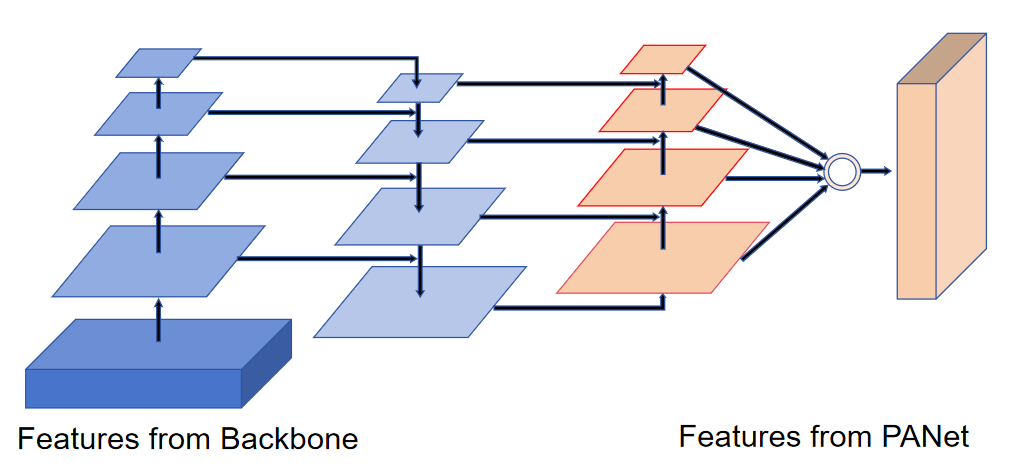}  
  \caption{The $PANet$  framework, based on $FPN$, adds a bottom-up path and employs feature concatenation. }
  \label{fig:pan}\vspace{-5mm}    
\end{figure}

The input to the model consists of images captured from the overhead roadside cameras. Initially, features are extracted from each image using a conventional Convolutional Neural Network (CNN) backbone~\cite{he2016resnet}. Subsequently, multi-scale features obtained from various stages of the CNN are fed into a Feature Pyramid Network (FPN)~\cite{tan2020efficientdet} to integrate comprehensive semantic information.  In this framework, we employ PANet~\cite{panet}, which enhances the fusion of low-level detail and high-level semantic information by incorporating a bottom-up pathway. This structure is particularly beneficial for improving detection capabilities for small and occluded objects, while also accommodating targets of varying scales, as illustrated in Fig.~\ref{fig:pan}. Finally, the pyramid features are upsampled to a uniform resolution and stacked, yielding the final output features.

\subsection{Map Generation Network}

In contemporary end-to-end mapping frameworks, the map generation module frequently utilizes traditional BEVFormer~\cite{li2203bevformer} architecture, which relies on camera intrinsic and extrinsic parameters to transform image features into the BEV feature space. This transformation is followed by a series of encoding and decoding processes to produce the feature map. The transformation from perspective view to BEV feature space typically employs techniques such as CVT~\cite{zhou2022cross}, LSS~\cite{philion2020lift}, GKT~\cite{chen2022efficient}, and IPM~\cite{mallot1991inverse}. Given the high complexity associated with the current BEVFormer architecture and the specific characteristics of the Navinfo dataset, we have adopted a BEV-Learner approach. Instead of the conventional BEVFormer pipeline, we implement a simplified Spatial Transformer Network (STN)~\cite{jaderberg2015spatial} alongside a basic CNN feature fusion network to encode image features, then decode these features to construct the map generation network and effectively complete the map generation task. The architecture consists of three parts: the Spatial Transformer Network (STN), the Feature Fusion Network, and the Map Decoder.

\subsubsection{Spatial Transformer Network (STN)~\cite{jaderberg2015spatial}} 
\begin{figure}
    \centering
  \includegraphics[width=0.5\textwidth]{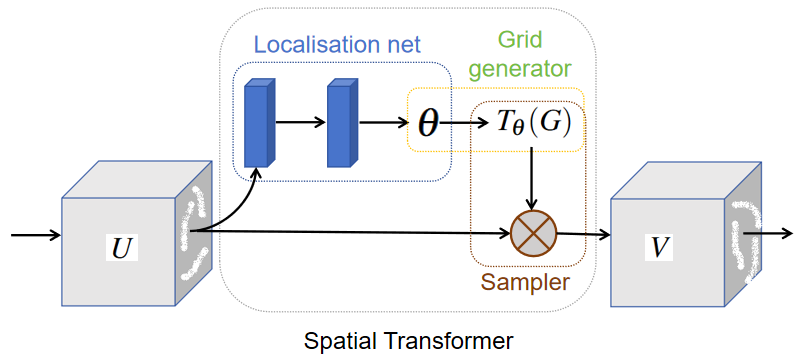}
  
  \caption{The architecture of a spatial transformer module. The input feature map $U$ is passed to a localisation network which regresses the transformation parameters $\theta$. The regular spatial grid $G$ over $V$ is transformed to the sampling grid $T_{\theta}(G)$, which is applied to $U$ and producing the warped output feature map $V$.}
  \label{fig:stn}\vspace{-5mm} 
\end{figure}

STN enhances the network's adaptability to geometric transformations. The fundamental principle behind STN is to integrate a learnable module within the standard CNN architecture, enabling explicit spatial transformations of the input image (or features). This capability ultimately improves the network's ability to accommodate geometric deformations present in the input data. The STN is composed of three components: the localization network, the grid generator, and the sampler, as depicted in Fig.~\ref{fig:stn}. We design an STN for each input image, with the corresponding features being processed through the localization network. The grid generator and sampler then conduct geometric transformations on the image features, thereby aligning them more closely with the BEV space features.

\subsubsection{Feature Fusion Network} This step is designed to amalgamate image features originating from different perspectives. Given that the proposed approach does not utilize camera parameters to map BEV space features but instead relies on geometric transformations derived from the STN—trained without explicit guidance—we implement a two-layer CNN featuring larger kernels to accomplish feature fusion. This design seeks to mitigate potential accuracy loss incurred by the prior operations, achieved by expanding the receptive field. The limited number of CNN layers is intentionally kept low to enhance computational efficiency, as the inclusion of additional layers could adversely affect performance.

\subsubsection{Map Decoder~\cite{maptr}} The hierarchical query embedding scheme of MapTR is continued to be used, and multiple cascaded decoding layers are employed to implement instance-level queries and point-level queries shared by all instances. This process retrieves the point set information for each map element. Subsequently, a classification branch and a point regression branch are used to process the point set. The classification branch predicts the instance class scores, while the point regression branch predicts the positions of the point set. For each map element, it outputs a \(2N_e\) dimensional vector representing the normalized BEV coordinates of the \(N_e\) points.

\subsection{Training Loss Functions}

Following the methodology outlined in hierarchical matching~\cite{maptr}, we define \(y= (c,V,\Gamma)\) to represent a ground truth map element, where \(c\) denotes the class, \(V\) is the point set, and \(\Gamma\) is the permutation group~\cite{maptr}. In contrast, we denote the predicted map element as \(\hat{y}= (\hat{p},\hat{V})\) with \(\hat{p}\) representing the class classification score and \(\hat{V}\) representing the predicted point set. For a given ground truth map element \(y_i\), we can identify a corresponding predicted map element \(\hat{y}_{\hat{\pi}(i)}\) that minimizes the matching cost. Here, \(\hat{\pi}(i)\) indicates the index of \(\hat{y}_{\pi(i)}\) in the list of predicted map elements, and it can be determined using the following formula:

\begin{equation}
\hat{\pi}(i) = \arg\min_{j \in \Pi_M} L_{\text{ins match}}(\hat{y}_{j}, y_i)
\end{equation}
where \(M\)   represents the total number of predicted map elements, and \(\Pi_M\) is defined as \(\{0, 1, 2, ..., M\}\).  The matching cost \(L_{ins\ match}(\hat{y}_{j}, y_i)\) between the predicted value \(\hat{y}_{j}\) and the ground truth value \(y_i\) can be expressed as:

\begin{equation}
L_{\text{ins match}}(\hat{y}_j, y_i) = L_{\text{FocalLoss}}(\hat{p}_j, c_i) + L_{\text{position}}(\hat{V}_j, V_i,\Gamma_{\gamma(j,i)})
\end{equation}

In this expression, \(\hat{p}_j\) denotes the class score of the \(j\)-th predicted map element, \(c_i\) denotes the class label of the \(i\)-th ground truth map element, and \(L_{FocalLoss}(\hat{p}_j, c_i)\) is the Focal Loss~\cite{lin2017focal} calculated for these values, functioning as the class matching cost term. The term \(L_{position}(\hat{V}_j, V_i, \Gamma_{\gamma(j,i)})\) serves as the position matching cost, where \(\hat{V}_j\) signifies the point set of the \(j\)-th predicted map element, \(V_i\) denotes the point set of the \(i\)-th ground truth map element, and \(\Gamma_{\gamma(j,i)}\) corresponds to the point set yielding the lowest matching cost found within the permutation group, utilizing the Manhattan distance. The function \(\gamma(j,i)\) is defined as:

\begin{equation}
\gamma(j,i) = \arg\min_{k \in \Pi_K} D_{\text{Manhattan}}(\hat{V}_{j}, V_{i},\Gamma_{k})
\end{equation}
where \(K\) represents the number of permutations in the permutation group of the point set, and \(\Pi_K\) represents \(\{0, 1, 2, ..., K\}\).

Based on the preceding definitions and reasoning, we establish that \(\hat{y}_{\hat{\pi}(i)}\) denotes the predicted result for the ground truth map element \(y_i\). The overall loss function comprises three components: classification loss, point-to-point loss, and edge direction loss~\cite{maptr}, expressed as follows:

\begin{equation}
{L} = \alpha_c {L}_{\text{cls}} + \alpha_p {L}_{\text{p2p}} + \alpha_d {L}_{\text{dir}}
\end{equation}

In this equation, \(\alpha_c\), \(\alpha_p\) and \(\alpha_d\) are the weights assigned to balance the different loss terms, while \(L_{cls}\), \(L_{p2p}\), and \(L_{dir}\) correspond to the classification loss, point-to-point loss, and edge direction loss, respectively.

\textbf{Classification Loss.} 
Utilizing the definition of \(L_{\text{FocalLoss}}\) provided earlier, the classification loss \(L_{cls}\) is defined  as follows:
\begin{equation}
{L}_{\text{cls}} = \sum_{i=0}^{N-1} {L}_{\text{FocalLoss}}\left(\hat{p}_{\hat{\pi}(i)}, c_i\right).
\end{equation}
where \(N\) denotes the total number of ground truth map elements.

\textbf{Point2point Loss.} The point-to-point loss supervises the position of each predicted point. Based on the definition of \(L_{\text{position}}\) outlined previously, the point-to-point loss \(L_{p2p}\) is defined as follows:
\begin{equation}
{L}_{\text{p2p}}  = \sum_{i=0}^{N-1} L_{\text{position}}(\hat{V}_{\hat{\pi}(i)}, V_i,\Gamma_{\gamma({\hat{\pi}(i)},i)})
\end{equation}
where \(N\) denotes the total number of ground truth map elements.

\textbf{Edge Direction Loss.} The point-to-point loss focuses solely on the node points of polylines and polygons, omitting considerations for edges (the connecting lines between adjacent points). To accurately represent map elements, the orientation of the edges is critical. Consequently, we have fully integrated the edge direction loss from the method employed in MapTR~\cite{maptr}, and further elaboration on this aspect will not be provided here.

\begin{table*}[htp!]
    \centering
    \begin{threeparttable}
    \caption{Comparison of different methods on the Navinfo dataset.}
    
    \label{tab:0}
    \begin{tabular}{lccccccccc}
        \toprule
        Method & Modality & Backbone & Epochs & InputImageNum & $AP_{ped}$ & $AP_{divider}$ & $AP_{boundary}$ & mAP & FPS \\        
        \midrule
        HDMapNet & C & ResNet50 & 110 & 4 & 31.3 & 43.5 & 38.5 & 37.8 & 0.6 \\
        \midrule
        VectorMapNet & C & ResNet50 & 110 & 4 & 34.3 & 50.5 & 37.2 & 40.7 & 3.8 \\
        \midrule
        MapTR (nano) & C & ResNet18 & 110 & 4 & 37.3 & 54.5 & 49.2 & 47.0 & 27.3 \\
        MapTR (tiny) & C & ResNet50 & 24 & 4 & 45.2 & 53.7 & 54.3 & 51.1 & 14.2 \\
        MapTR (tiny) & C & ResNet50 & 110 & 4 & 55.7 & 61.9 & 61.7 & 59.8 & 14.2 \\
        \midrule
        MRC-VMap (nano) & C & ResNet18 & 110 & 4 & 39.1 & 56.6 & 51.3 & 49.0 & \textbf{40.9} \\
        MRC-VMap (tiny) & C & ResNet50 & 24 & 4 & 47.9 & 56.8 & 57.5 & 54.1 & 18.2 \\
        MRC-VMap (tiny) & C & ResNet50 & 110 & 4 & 58.7 &  64.8 & 65.5 & \textbf{63.0} & 18.2 \\
        \midrule
        *VI-Map & C\&L & ResNet50 & \-- & 4 & 60.1 & 66.2 & 66.6 & \textbf{64.3} & \-- \\        
        \bottomrule
    \end{tabular}
    \begin{tablenotes}
    
    \scriptsize
     \item {Table~\ref{tab:0} presents a comparison of our method with state-of-the-art approaches (~\cite{li2022hdmapnet}, ~\cite{liu2023vectormapnet}, ~\cite{maptr}, ~\cite{vi-map}) utilizing the Navinfo validation set. The designations "C" and "L" correspond to camera-based and LiDAR-based modalities, respectively. The Average Precision (AP) and mean Average Precision (mAP) values reported were derived from our experiments, while the Frames Per Second (FPS) metrics were measured on a consistent hardware platform equipped with an RTX 3090 GPU. The quantitative findings demonstrate that under the camera-based modality, MRC-VMap (tiny) significantly outperforms the other methods listed in the table. MRC-VMap (nano) achieves state-of-the-art performance in the camera-based category, running at an impressive 40.9 FPS. 
     *VI-Map: For the VI-Map method, tests were conducted on the Navinfo dataset using modality different from other methods (camera-based and LiDAR-based).}
    \end{tablenotes}
    \end{threeparttable}
    \vspace{-2mm}
\end{table*}

\begin{table*}[htp!]
    \centering
    \begin{threeparttable}
    \caption{Comparison of different methods on the NuScenes dataset.}
    
    \label{tab:1}
    \begin{tabular}{lccccccccc}
        \toprule
        Method & Modality & Backbone & Epochs & InputImageNum & $AP_{ped}$ & $AP_{divider}$ & $AP_{boundary}$ & mAP & FPS \\      
        \midrule
        VectorMapNet & C & ResNet50 & 110 & 6 & 36.1 & 47.3 & 39.3 & 40.9 & 2.9 \\
        \midrule
        MapTR (nano) & C & ResNet18 & 110 & 6 & 39.6 & 49.9 & 48.2 & 45.9 & 25.1 \\
        MapTR (tiny) & C & ResNet50 & 24 & 6 & 46.3 & 51.5 & 53.1 & 50.3 & 11.2 \\
        MapTR (tiny) & C & ResNet50 & 110 & 6 & 56.2 & 59.8 & 60.1 & 58.7 & 11.2 \\
        \midrule
        MRC-VMap (nano) & C & ResNet18 & 110 & 6 & 39.8 & 50.4 & 48.4 & 46.2 & \textbf{38.1} \\
        MRC-VMap (tiny) & C & ResNet50 & 24 & 6 & 46.2 & 51.7 & 53.6 & 50.5 & 14.7 \\
        MRC-VMap (tiny) & C & ResNet50 & 110 & 6 & 58.1 &  62.5 & 61.2 & \textbf{60.6} & 14.7 \\    
        \bottomrule
    \end{tabular}
    \begin{tablenotes}
    
    \scriptsize
     \item {Table~\ref{tab:1} presents a comparison of our method with state-of-the-art approaches (~\cite{liu2023vectormapnet}, ~\cite{maptr}) utilizing the NuScenes validation set. The definitions of the fields in this table are identical to those in Table ~\ref{tab:0}. The quantitative findings demonstrate that MRC-VMap (tiny) significantly outperforms the other methods listed in the table. MRC-VMap (nano) achieves state-of-the-art performance, running at an impressive 38.1 FPS.}
    \end{tablenotes}
    \end{threeparttable}
    \vspace{-5mm}
\end{table*}

\section{Experiments}
\subsection{Real-world Dataset}

In this study, we developed a custom dataset to facilitate our research objectives and named Navinfo dataset, as depicted in Fig.~\ref{fig:navinfo}. The dataset is primarily comprised of video surveillance data from traffic intersections. Each intersection is monitored by cameras positioned to capture images from four distinct orientations. Each camera boasts a resolution of 2 million pixels, allowing for high-quality imagery. The combined view from these four cameras provides comprehensive coverage of the entire intersection as well as a portion of the roadways.

In addition to image information, Navinfo dataset also encompasses several important information. Firstly, the camera metadata includes intrinsic parameters, quaternions, Euler angles (pitch, yaw, roll – these are labeled once and may vary from actual values due to the long data collection period), geographical coordinates (latitude and longitude of WGS84 coordinate system~\cite{kelly2022transforming}), elevation, and real-world coordinates relative to the camera's position. Secondly, the map information includes detailed descriptions of ground elements such as lane markings, road edges, pedestrian crossings, lane dividers, road boundaries, directional arrows, inscriptions on the road surface, ground signs, and traffic islands. Additionally, elevation elements such as traffic lights, signage, and guardrails are included, all with precise geographic coordinates and shape information. By applying camera parameters to process the images, we establish a one-to-one correspondence with the map information.

In terms of Navinfo dataset scale, data collection was conducted across 4,000 intersections in four major cities: 1,246 intersections in Beijing, 1,121 in Shanghai, 988 in Guangzhou, and 645 in Shenzhen. A total of 16,000 roadside camera datasets were compiled, which include 160,000 one-minute video clips captured periodically every 25 hours per camera, all of which were recorded during daytime conditions. This dataset enables the generation of 400,000 image groups, amounting to a total of 1.6 million images, alongside 16,000 groups of ground truth data. Importantly, the map information (ground truth) was meticulously collected by NavInfo using a laser-equipped vehicle, thereby ensuring a high degree of precision and reliability in the dataset.

\begin{figure*}
\centering
  \includegraphics[width=0.9\textwidth]{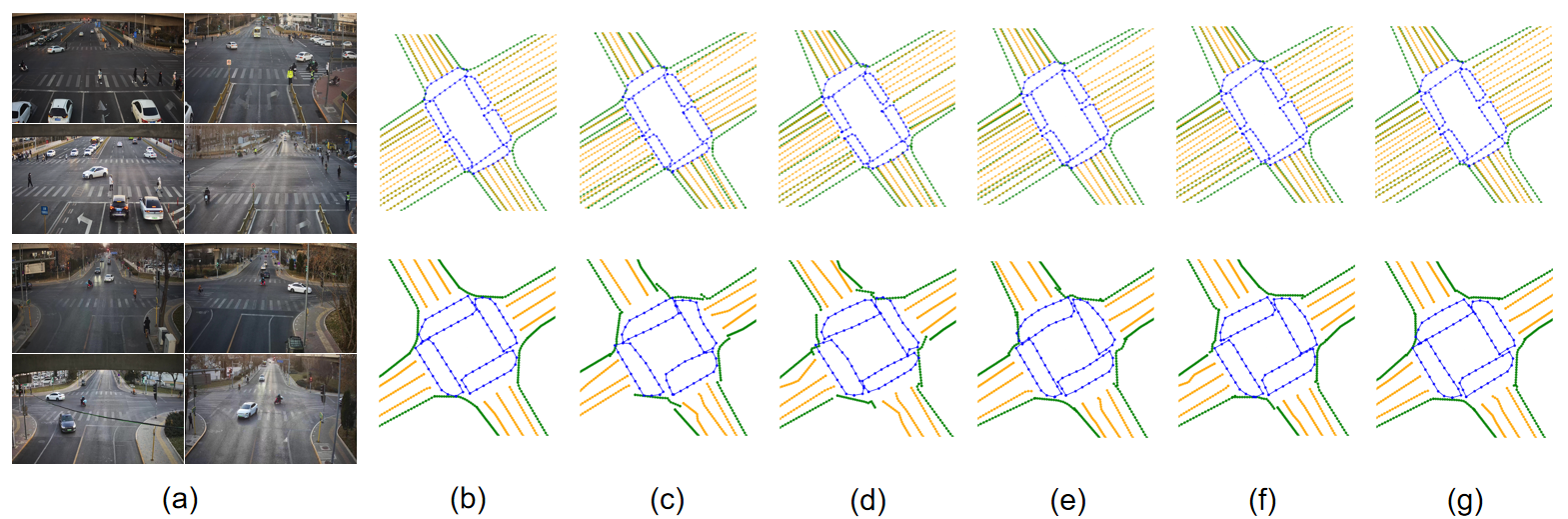}
  \caption{Comparison of our method with several existing methods on the Navinfo dataset, where (a) is the scene image, (b) is the ground truth of the corresponding intersection, (c) is the prediction result of HDMapNet, (d) is the prediction result of VectorMapNet, (e) is the prediction result of MapTR, (f) is the prediction result of MRC-VMap, and (g) is the prediction result of VI-Map (Camera-based and LiDAR-based). In these results, dividers, boundaries, and pedestrian crossings are represented in yellow, green, and blue, respectively.}
  \label{fig:real_gt_pd}
      \vspace{-5mm}
\end{figure*}

\subsection{Evaluation Metrics}

We have verified the effectiveness of the algorithm on the Navinfo dataset, which is similar to the popular NuScenes~\cite{caesar2020nuscenes} dataset format. Following the validation method of MapTR on the NuScenes dataset, we selected three types of map elements for a fair evaluation — pedestrian crossings, lane dividers, and road boundaries. The perception range of the NuScenes dataset is set to \([-15.0m, 15.0m]\) on the \(X\)-axis and \([-30.0m, 30.0m]\) on the Y-axis. Due to the different scenes of the Navinfo and NuScenes datasets, we have changed the perception range, adjusting it to \([-30.0m, 30.0m]\) on the \(X\)-axis, and the \(Y\)-axis is not changed. We use average precision (AP) and mean average precision (mAP) to assess the quality of map construction. The Chamfer distance \(D_{\text{Chamfer}}\) is utilized to determine whether the predictions match the ground truth (GT). We calculate \(AP_\tau\) under several \(D_{\text{Chamfer}}\) thresholds (\(\tau \in T, T = \{0.5, 1.0, 1.5\}\)), and then take the average across all thresholds as the final AP metric:

\begin{equation}
AP = \frac{1}{|T|} \sum_{\tau \in T} AP_{\tau}.
\end{equation}

mAP represents the average of AP values for various elements. In this study, we evaluated three specific elements: pedestrian crossings, lane dividers, and road boundaries, which are denoted as \(AP_\textit{ped}\), \(AP_\textit{divider}\), and \(AP_\textit{boundary}\), respectively. Accordingly, the value of mAP can be expressed as follows:

\begin{equation}
mAP = \frac{AP_\textit{ped}+AP_\textit{divider}+AP_\textit{boundary}}{3}
\end{equation}
where \( 3 \) signifies that there are three types of elements.

\subsection{Edge Deployment}

This project aims to implement a range of algorithms on edge devices, selecting the Jetson AGX Orin as the target platform. It is compatible with NVIDIA JetPack SDK. The process begins with configuring the JetPack SDK to access CUDA, cuDNN, and TensorRT, and setting up the GCC compilation environment. The algorithm model was converted to a TensorRT-compatible format and optimized to FP16 precision to maintain accuracy while improving efficiency. Finally, the implementation involved data reading, model preprocessing, and postprocessing in C++, with visual results rendered based on the postprocessing outcomes.

Upon deployment, The MRC-VMap, which uses ResNet50 as its backbone network,  occupies approximately 1.6GB of GPU memory and achieves an inference speed of 20 FPS, while maintaining an overall accuracy degradation of less than 1\%.

\subsection{Comparison Results}
This research represents the initial effort to investigate the generation of high-definition maps utilizing multi-camera data from intersections, leading to a scarcity of widely recognized comparative methods in the existing literature. However, given that the data structure of the Navinfo dataset is modeled after the well-established NuScenes dataset format, we are able to conduct comparative experiments on the Navinfo dataset by assessing several classic open-source algorithms, such as HDMapNet. This comparison follows the algorithm performance evaluation approach employed in studies utilizing the NuScenes dataset.

Table~\ref{tab:0} summarizes the experimental results from various algorithms, including HDMapNet, VectorMapNet, MapTR, MRC-VMap, and VI-Map, on the Navinfo dataset. The comparison primarily focuses on the accuracy of three key elements: pedestrian crossings, lane dividers, and road boundaries. The results indicate that the MRC-VMap significantly improves the quality of high-definition  map generation, achieving an increase of over 3 mAP in pure visual mode. This outcome underscores the effectiveness of the MRC-VMap algorithm in generating high-quality, high-definition maps within the framework of the Navinfo dataset.

Furthermore, we assessed the efficiency of MRC-VMap, finding it to be 28\% more efficient than the current leading algorithm, MapTR, when executed on an RTX 3090 GPU. Lastly, we implemented the VI-Map method on the Navinfo dataset, utilizing both camera and LiDAR point cloud data as inputs; its accuracy was reported to be merely 1.3 mAP higher than that of MRC-VMap. The prediction results of the vectorized high-definition maps are illustrated in Fig.~\ref{fig:real_gt_pd}.

Table~\ref{tab:1} summarizes the experimental results from various algorithms, including VectorMapNet, MapTR and MRC-VMap on the NuScenes dataset. The results show that MRC-VMap achieves a 1.9 mAP increase and its operational efficiency has improved by 31\% compared to MapTR. This also indicates that MRC-VMap can enhance the quality and efficiency of high-definition map generation.

\subsection{Ablation Study}

This section evaluates the effectiveness of the proposed modules and design choices through comprehensive ablation studies. To ensure fairness in comparison, all experiments were conducted on the Navinfo dataset over a training schedule of 110 epochs, utilizing ResNet18 and ResNet50 as the backbone networks for image feature extraction.

\begin{table}[htp!]
    \centering    
    \caption{Performance comparison of different backbones and FPN methods.}
    \label{tab:2}
    \begin{tabular}{lccc}
        \toprule
        Backbone & Method & mAP &  FPS \\
        \midrule
        \multirow{2}{*}{ResNet18} & FPN & 47.6   &  42.1\\
        & PANet & 49.0   & 40.9 \\
        \midrule
        \multirow{2}{*}{ResNet50} & FPN & 61.0   &  18.8\\
        & PANet & 63.0   & 18.2 \\
        \bottomrule
    \end{tabular} 
\end{table}

\subsubsection{Ablation on FPN Variants} Table~\ref{tab:2} demonstrates the effectiveness of two variants of Feature Pyramid Networks within the image feature extraction module. Regardless of whether ResNet18 or ResNet50 is employed, the PANet consistently yields superior accuracy. Specifically, when utilizing ResNet18, the accuracy of the algorithm increases by 1.4 mAP, albeit with a slight reduction in efficiency (a drop of 8\% in frames per second). In the case of ResNet50, the accuracy improves by 2 mAP while maintaining nearly the same efficiency (frames per second remain unchanged). These findings suggest that employing PANet as the neck network for image feature extraction effectively enhances the algorithm's accuracy without compromising efficiency when the backbone is ResNet50.

\begin{table}[htp!]
    \centering
    \caption{Performance comparison of different backbones and BEV acquisition methods.}
    \label{tab:3}
    \begin{tabular}{lccc}
        \toprule
        Backbone & Method & mAP &  FPS \\
        \midrule
        \multirow{3}{*}{ResNet18} & GKT & 49.1 &  26.1 \\
        & LSS & 48.3 &  25.6\\
        & BEV-Learner & 49.0 & 40.9 \\
        \midrule
        \multirow{3}{*}{ResNet50} & GKT & 63.2 &  13.8 \\
        & LSS & 63.1 &  13.4 \\
        & BEV-Learner & 63.0 &  18.2 \\
        \bottomrule
    \end{tabular}   
    \vspace{-2mm}
\end{table}

\subsubsection{Ablation on BEV Extractor} Evaluating various methodologies for transforming perspective view image features into the BEV space features, Table~\ref{tab:3} presents the results for methods including LSS, GKT, and the proposed BEV-Learner approach. The experimental outcomes demonstrate that the BEV-Learner method, regardless of whether ResNet18 or ResNet50 is used, consistently exhibits higher operational efficiency while achieving accuracy levels comparable to the current best-performing GKT method. Specifically, the implementation of the BEV-Learner method led to a 57\% improvement in efficiency when utilizing ResNet18, and a 32\% enhancement when using ResNet50, all while maintaining accuracy on par with the GKT approach.

\begin{table}[htp!]

    \centering
    \caption{Performance comparison of different quantities of inputs.}
    \label{tab:4}
    \begin{tabular}{lccc}
        \toprule
        Mode & Backbone & mAP &  FPS \\
        \midrule
        MRC-VMap (S) & ResNet50 & 47.9 & 33.6  \\
        
        \midrule
        MRC-VMap (M) &  ResNet50 & 63.0 & 18.2  \\
        \bottomrule
    \end{tabular}
\end{table}

\subsubsection{Ablation Study on Single-Camera Mode} Table~\ref{tab:4} exhibits the performance of MRC-VMap when utilizing data from a single camera as input in an intersection scenario. The analysis reveals that when ResNet50 serves as the backbone of the algorithm, the single-camera mode presents certain advantages in terms of efficiency when compared to the four-camera mode (as detailed in Table~\ref{tab:1}). However, it is noteworthy that there remains a substantial discrepancy in accuracy, with a reduction of 15.1 mAP evident in the single-camera configuration.

\section{Conclusion}
This study introduced MRC-VMap, a vision-centric, cost-effective framework for the generation of high-precision vector maps in autonomous driving. By leveraging existing roadside cameras and an end-to-end neural architecture, our approach addresses the high costs of LiDAR-based mapping and overcomes limitations of online methods, such as restricted views. Optimized for commodity-embedded AI chips, such as the Jetson AGX Orin, MRC-VMap fuses time-synchronized images from multiple angles to produce complete intersection maps in real time. Extensive tests on 4,000 intersections in four major Chinese cities show that MRC-VMap achieves an mAP of 63.0 and operates at 40.9 FPS, with a 28\% efficiency improvement over leading methods, demonstrating its scalability and practicality for real-world deployment.

%This paper proposes a vision-centric framework for vectorized map construction at intersections by roadside cameras. We formulate the framework as an generative neural network (i.e., MRC-VMap), which mainly takes time-aligned surveillance photos from multiple roadside cameras of a intersection as inputs, and directly outputs a vectorized map for the intersection. In addition, the neural architecture of MRC-VMap is designed to be deployed on embedded AI chips in roadside edge devices. Through end-to-end learning, MRC-VMap can produce high-definition maps that are closest to expert mapping, reducing errors introduced by intermediate modules such as feature extraction and BEV conversion. More importantly, roadside cameras from multiple directions can significantly improve the completeness of vectorized mapping and alleviate occlusions. We conducted extensive experiments involving $4000$ intersections of $4$ metropolitan areas in China. Results showed that MRC-VMap achieved $63.0$ higher mAP than the state-of-the-art online method and comparable performance to LiDAR-based offline approaches. 

% \section*{Acknowledgments}
% This work was sponsored by XXX. 

\balance
\bibliographystyle{IEEEtran}
\bibliography{IEEEexample}
\end{document}